%% file: main.tex
\documentclass[10pt,twocolumn,letterpaper]{article}

\usepackage{cvpr}              % To produce the CAMERA-READY version
\usepackage{graphicx}
\usepackage{amsmath}
\usepackage{amssymb}
\usepackage{booktabs}
\usepackage{multirow} 
\usepackage{tabularx}
\usepackage{algorithm}
\usepackage{algpseudocode}
\usepackage{colortbl}      % Provides tools to color rows and cells in tables
\usepackage[T1]{fontenc}
\usepackage{threeparttable} % To use footnotes in tables
\usepackage[absolute,overlay]{textpos}
\definecolor{cvprblue}{rgb}{0.21,0.49,0.74}
\usepackage[pagebackref,breaklinks,colorlinks,allcolors=cvprblue]{hyperref}

\def\paperID{7245} % *** Enter the Paper ID here
\def\confName{CVPR}
\def\confYear{2025}

\title{LHPF: Look back the History and Plan for the Future in Autonomous Driving}
\author{Sheng Wang\textsuperscript{1}, Yao Tian\textsuperscript{1}, Xiaodong Mei\textsuperscript{1}, Ge Sun\textsuperscript{1}, Jie Cheng\textsuperscript{1}, Fulong Ma\textsuperscript{2} \\Pedro V. Sander\textsuperscript{1}, Junwei Liang\textsuperscript{1,2,†}\\
\textsuperscript{1}HKUST ~\  \textsuperscript{2}HKUST(GZ)\\
{\tt\small †Corresponding Author} \\ {\tt\small swangei@connect.ust.hk  junweiliang@hkust-gz.edu.cn} \\
{\tt\small \url{https://chantsss.github.io/LHPF/}}
\vspace{-0.5cm} 
}

\begin{document}
\maketitle 
\input{sec/0_abstract}
\vspace{-0.5cm}  
\input{sec/1_intro}

\input{sec/2_related}
\input{sec/3_method}

\input{sec/4_experiments}
\input{sec/5_conclusion}

\clearpage
{
    \small
    \bibliographystyle{ieeenat_fullname}
    \bibliography{main}
}
\clearpage
\input{sec/6_supplementary}


\end{document}

%% file: sec/0_abstract.tex
\begin{abstract}
Decision-making and planning in autonomous driving critically reflect the safety of the system, making effective planning imperative. Current imitation learning-based planning algorithms often merge historical trajectories with present observations to predict future candidate paths. However, these algorithms typically assess the current and historical plans independently, leading to discontinuities in driving intentions and an accumulation of errors with each step in a discontinuous plan. To tackle this challenge, this paper introduces LHPF, an imitation learning planner that integrates historical planning information. Our approach employs a historical intention aggregation module that pools historical planning intentions, which are then combined with a spatial query vector to decode the final planning trajectory. Furthermore, we incorporate a comfort auxiliary task to enhance the human-like quality of the driving behavior. Extensive experiments using both real-world and synthetic data demonstrate that LHPF not only surpasses existing advanced learning-based planners in planning performance but also marks the first instance of a purely learning-based planner outperforming the expert. Additionally, the application of the historical intention aggregation module across various backbones highlights the considerable potential of the proposed method. The code will be made publicly available.
\end{abstract}

%% file: sec/1_intro.tex
\vspace{-0.2cm}
\section{Introduction}
\vspace{-0.1cm}
\label{sec:intro}

\begin{figure}[t]
  \centering
   \includegraphics[width=0.95\linewidth]{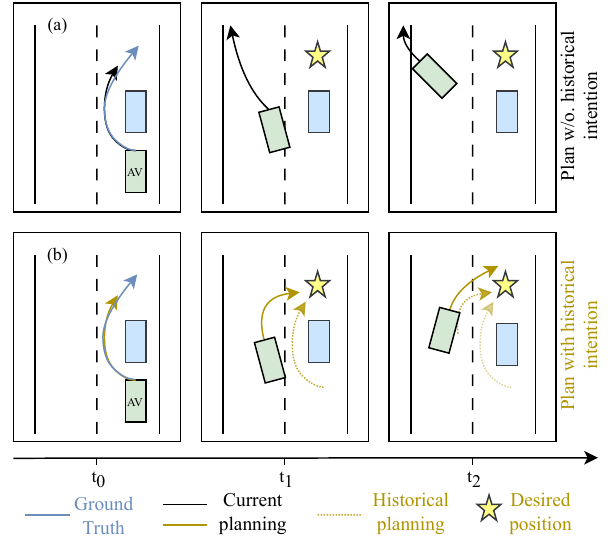}
   \vspace{-0.2cm}
\caption{Comparison of imitation learning planners with and without historical intentions. The ego vehicle, shown as the green rectangle, initiates an overtaking maneuver at \( t_0 \). Both planners in (a) and (b) exhibit high open-loop performance at \( t_0 \); however, due to the absence of historical intentions and the accumulation of errors, method (a) diverges significantly from the ground truth by \( t_2 \). In contrast, the planner in (b) incorporates historical planning embeddings, preserving the initial driving intention accurately.}
   \label{fig:intro}
\vspace{-0.7cm}
\end{figure}

Effective decision-making and strategic planning are essential for autonomous driving, ensuring system safety. While rule-based methods are reliable, they face scalability issues and require substantial resources to adapt to new scenarios. In contrast, learning-based methods \cite{plantf, mpnp, gameformer, UniAD, transfuser, PDM} offer better scalability and adaptability with reduced human intervention. The emergence of benchmarks like the NuPlan challenge \cite{nuplan} has shifted the focus toward methods capable of handling dynamic, unpredictable environments.

Imitation learning (IL)-based approaches, which replicate human driving policies, have become a cornerstone of autonomous driving. These planners convert observation data from perception modules into rasterized images \cite{nuplan, learning_by_cheating, exploring_limitations_of_bc} or vectorized graph structures \cite{UrbanDriver, GC-PGP, PDM}. By leveraging large-scale, cost-effective data, these methods reduce reliance on expensive labeled datasets.  Despite the advancements of IL-based methods in open-loop performance, they still struggle in closed-loop settings. Specifically, in real-world driving, accumulated errors can cause significant deviations from the training distribution. This issue arises because current IL-based planners generate plans based solely on the current observation and ego states. Once errors have accumulated, the planner, unable to recognize previous planning intentions, cannot correct the deviations, leading to further divergence from the intended driving path.
For example, as shown in Fig. \ref{fig:intro}(a), an IL planner with intention close to the ground truth at \( t_0 \), which exacerbates an ego vehicle's left lane cut during an overtaking maneuver. However, due to the absence of historical intentions and the accumulation of errors, the planning result diverges significantly from the ground truth by \( t_2 \). Recently, motion prediction methods \cite{hpnet}, \cite{dcms}, \cite{tesla} have suggested using time-scale signals and multi-frame integration to improve prediction accuracy. This has inspired us to explore the impact of temporal factors on the closed-loop performance of IL-based planners. 
Due to the high initial prediction accuracy exhibited by IL-based planners, incorporating historical planning intentions into the current process provides an effective way to maintain temporal consistency and correlations, as shown in Fig. \ref{fig:intro}(b).
% This exploration is particularly valuable because, due to advancements in prediction tasks, imitation learning-based planners often exhibit high initial prediction accuracy. Incorporating the initial planning intentions into the current planning process provides an effective way to correct accumulated errors as shown in Fig. \ref{fig:intro}(b). 

Building on these insights, we introduce LHPF, a novel imitation learning planner that integrates historical planning information. Unlike conventional methods that focus on current planning, LHPF processes multiple consecutive frames to capture driving behaviors over time. This approach improves performance by merging historical intentions with spatial embeddings, which are processed by a query-based decoder to generate the final trajectory. Comfort-related tasks are also introduced to promote human-like driving. Our contributions are threefold: \textbf{First}, we propose a Spatio-Temporal aggregation and query-based decoding module, with a comfort auxiliary task, integrating temporal and spatial consistency for a deeper understanding of driving intentions, enabling more context-aware decision-making. \textbf{Second}, our planner achieves the highest performance among purely learning-based methods on the real-world dataset, with a reactive score of 81.25. To the best of our knowledge, this is the first purely learning-based method to surpass expert performance, marking a significant milestone in autonomous driving. \textbf{Third}, experiments on different datasets and backbones show that LHPF can serve as a plug-in, enhancing existing IL-based planner without modifying the backbone and requiring only five fine-tuning epochs, making it adaptable to various systems.

%% file: sec/2_related.tex
\vspace{-0.2cm}
\section{Related Work}
\vspace{-0.2cm}
\noindent\textbf{Imitation Learning for Planning.} Imitation learning (IL) has become a key approach in autonomous driving, especially through end-to-end (E2E) methods that replicate human driving policies. These methods leverage large-scale, cost-effective data, eliminating the need for expensive labeled datasets. Early approaches relied on CNN-based models to process camera inputs for driving guidance \cite{learning_by_cheating, exploring_limitations_of_bc}. However, focus has shifted to advanced techniques, such as multi-sensor fusion \cite{transfuser, neat, Safety-enhanced} and modular E2E architectures \cite{LAV, UniAD}, which integrate perception, prediction, and planning \cite{vad, stp3}. Despite these advances, many methods rely on still suffer from the sim-to-real problem. specifically, this challenge leads to issues such as covariate shift and cumulative errors when transferring learned behaviors to real-world driving dynamics. Some prior works have addressed this by using a mid-to-mid approach with real post-perception data, successfully demonstrated in projects like ChauffeurNet \cite{chauffeurnet}, SafetyNet \cite{safetynet}, and UrbanDriver \cite{urban_driver}. Notable methods include PDM-Hybrid \cite{PDM}, which refines trajectories with rule-based center lines, and GameFormer \cite{gameformer}, which improves scene understanding with a hierarchical query-based attention mechanism. Other methods, such as Pluto \cite{pluto}, integrate auxiliary losses and strong prior information. PlanTF \cite{plantf} uses augmentation and dropout techniques to mitigate errors, while DTPP \cite{dtpp} combines ego-conditioned prediction with cost modeling via inverse reinforcement learning (IRL). A major development in these works is the use of vector-based models, improving predictive accuracy. We build on this by employing a novel query-based architecture \cite{pluto} that models both longitudinal and lateral driving behaviors
more flexibly and diversely. Additionally, we introduce an auxiliary loss focused on driving comfort, aiming to refine the
smoothness of learned behaviors and ensure consistent, safe
driving dynamics.

\noindent\textbf{Multimodal Trajectory Generation.} The unpredictable nature of vehicle driving behaviors necessitates the generation of multiple feasible trajectories under identical observational conditions. Capturing a diverse set of potential trajectories is critical for effective prediction and planning. The trajectory forecasting community has made significant contributions in this area, with extensive research exploring anchor-based methods to classify discrete intents and directly regress multiple future trajectories along with their associated probabilities \cite{multimodal_pre1, multimodal_pre2, multimodal_pre3, mpnp}. These efforts typically focus on target-based indicators, such as specific locations \cite{tnt, densetnt} or lanes \cite{multimodal_pre2}. State-of-the-art prediction and planning approaches, including those utilizing DETR-like architectures \cite{qcnet, hpnet, tesla, pluto}, tackle the multimodal challenge by combining anchor-free and anchor-based techniques, leading to substantial advancements. However, relying solely on spatial dimensions for mode classification can lead to discontinuous behavior between frames during closed-loop evaluation. Recent studies in trajectory prediction \cite{hpnet, dcms, tesla} have shown that information from trajectories generated between neighboring frames contains critical behavioral intention data on a temporal scale. Leveraging time-scale signals has enhanced open-loop prediction performance by incorporating historical data and multi-frame integration, improving both time consistency and prediction accuracy. Despite these technological advancements, the role of temporal factors in planning tasks remains underexplored, yet it is essential for ensuring the safety of autonomous driving systems.

%% file: sec/3_method.tex
\vspace{-0.4cm}
\section{Approach}
\begin{figure*}[ht]
    \centering
    \includegraphics[width=1\textwidth]{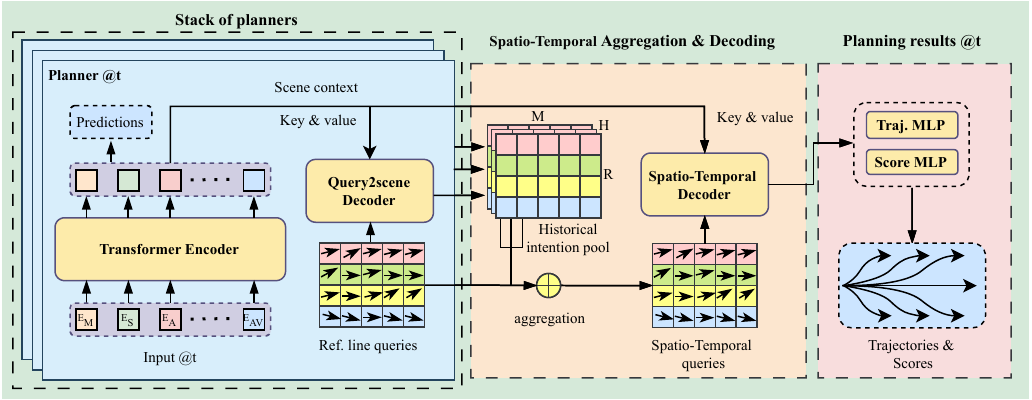}
    \vspace{-0.7cm}
    \caption{\textbf{Planning with Historical Intentions.} A stack of planners generates historical planning embeddings at each time step, which are stored in a historical intention pool and combined with reference line queries. Spatio-temporal queries are then aggregated and processed using self and cross attention with scene context as keys and values. Finally, the current planning embedding is passed through a multi-layer perceptron to generate future trajectories and scores.}

    \label{fig:overview_structure}
\vspace{-0.5cm}
\end{figure*}

\vspace{-0.2cm}
The proposed LHPF is an end-to-end autonomous driving framework consisting of a stack of planners at each timestamp. The overall architecture is shown in Fig. \ref{fig:overview_structure}. Each planner including an encoder to transform the postperception results into a representation vector, and a spatial query based decoder to generate future trajectories of the ego agent based on the representation vector. The latent representation, historical planning embeddings then stored in a historical intention pool. An aggregation between spatial query and temporal embeddings is employed to extract the spatio-temporal information for finally output of future planning trajectories at current step.

\subsection{Problem Formulation}

This work addresses autonomous driving trajectory planning in dynamic urban environments, aiming to generate safe and feasible trajectories for an autonomous vehicle (AV) that adapt to interactions with dynamic agents (e.g., vehicles, bicycles, pedestrians) and static obstacles. The input scene context consists of: \textbf{Dynamic agents} \( \mathcal{A} = \{\mathcal{A}_0, \mathcal{A}_1, \dots, \mathcal{A}_{N_A}\} \), where \( \mathcal{A}_0 \) represents the ego vehicle and \( \mathcal{A}_{1:N_A}\) are surrounding dynamic agents; \textbf{Static obstacles} \( \mathcal{S} = \{\mathcal{S}_1, \dots, \mathcal{S}_{N_S}\} \), representing non-traversable elements like barriers or curbs; and \textbf{Map and traffic information} \( \mathcal{M} \) and contextual data \( \mathcal{C} \), which include lane geometry, drivable areas, and traffic signals. Given past observations over a time window \(T_H\), the planner generates \( N_T \) candidate future trajectories \( \{\tau_1, \dots, \tau_{N_T}\} \), from which a scoring module selects the optimal trajectory \( \tau^{*} \) for the ego vehicle. The problem formulation is defined as:
\vspace{-0.2cm}
\[
(\tau^{*}, \pi^{*}), \mathcal{P}_{1:N_A} = G(\mathcal{A}, \mathcal{S}, \mathcal{M}, \mathcal{C} \mid \psi),
\vspace{-0.2cm}
\]
where \( G \) is the planning model with parameters \( \psi \), producing the optimal ego trajectory and its confidence score \( (\tau^{*}, \pi^{*}) \). Additionally, single-modal predictions \( \mathcal{P}_{1:N_A} \) for surrounding agents are generated to enhance the model’s comprehensive understanding of the dynamic environment.

% \subsection{Problem Formulation}
% In this study, we explore autonomous driving in dynamic urban environments, incorporating a diverse array of elements including an autonomous vehicle (AV), \(N_D\) dynamic agents, \(N_O\) static obstacles, a high-definition map \(M\), and a range of traffic-related contextual factors \(C\), such as the status of traffic lights. We define the feature set of agents as \(A = \{A_0, A_1, \ldots, A_{N_D}\}\), where \(A_0\) represents the AV and static obstacles are noted by \(O = \{O_1, O_2, \ldots, O_{N_O}\}\). The future state of each agent \(a\) at time \(t\) is denoted as \(y_t^a\), with history and future horizons represented by \(T_H\) and \(T_F\), respectively. LHPF is designed to concurrently generate \(N_T\) multimodal planning trajectories for the AV and predict the movements of each dynamic agent. The planning system is mathematically formulated as follows:
% \begin{align*}
% (T^{*}, \pi^{*}), P_{1:N_D} &= f(A, O, M, C \mid \theta),
% \end{align*}
% where the function \(f\) represents the neural network architecture of LHPF, \(\theta\) denotes the model parameters, and \((T^{*}, \pi^{*}) = \{(y_{0:T_F}^{i}, \pi_i) \mid i = 1 \ldots N_T\}\) encapsulates the planned trajectories and their corresponding confidence scores for the AV. Predictions for the agents are aggregated by \(P_{1:N_D} = \{y_{0:T_F}^a \mid a = 1 \ldots N_D\}\).

\subsection{Input Representation and Scene Encoding}

\noindent\textbf{Dynamic Encoding.} The state of each agent at time \(t\) is represented as \(s_t^i = (p_t^i, \theta_t^i, v_t^i, b_t^i, I_t^i)\), where \(p\) and \(\theta\) denote position and heading, \(v\) is the velocity, \(b\) represents the bounding box dimensions, and \(I\) is a binary mask for observation status. The agent’s historical state is represented as the difference between consecutive time steps: \(\hat{s}_t^i = (p_t^i - p_{t-1}^i, \theta_t^i - \theta_{t-1}^i, v_t^i - v_{t-1}^i, b_t^i, I_t^i)\), forming the feature vector \(F_A \in \mathbb{R}^{N_A \times (T_H-1) \times 8}\). A neighbor attention-based Feature Pyramid Network (FPN) is used to extract historical features, producing the agent embedding \(E_A \in \mathbb{R}^{N_A \times D}\), where \(D\) is the dimensionality of the hidden layers. The AV’s state is encoded with an attention-based state dropout encoder (SDE) \cite{plantf}, yielding the AV embedding \(E_{AV} \in \mathbb{R}^{1 \times D}\).

\noindent\textbf{Static Encoding.} Unlike motion forecasting tasks, where the map consists of \(N_M\) polylines, we compute feature vectors for each point in a polyline. Each point’s feature vector includes eight channels: \((p_i - p_0, p_i - p_{i-1}, p_i - p_{\text{left}}^i, p_i - p_{\text{right}}^i)\), where \(p_0\) is the polyline start, and \(p_{\text{left}}^i\) and \(p_{\text{right}}^i\) are the left and right boundaries. These features are vital for defining the drivable area. The polyline features are represented as \(F_P \in \mathbb{R}^{N_P \times n_p \times 8}\), processed using a PointNet-like encoder to obtain the polyline embeddings \(E_M \in \mathbb{R}^{N_M \times D}\). Static obstacles, such as traffic cones or barriers, are represented by \(o_i = (p_i, \theta_i, b_i)\) and encoded with a two-layer MLP to produce features \(F_S \in \mathbb{R}^{N_S \times 5}\), which are mapped to embeddings \(E_S \in \mathbb{R}^{N_S \times D}\).

\noindent\textbf{Scene Encoding.} To model interactions among different inputs, we concatenate the embeddings into a single tensor \(E_0 \in \mathbb{R}^{(N_A + N_S + N_M + 1) \times D}\), which is then processed by \(L_{\text{enc}}\) Transformer encoders. Global positional embeddings \(PE\), based on Fourier embeddings of the latest positions, and semantic attribute embeddings \(E_{\text{attr}}\) are added for enhanced spatial and contextual awareness. The initialization and transformation of \(E_0\) through the Transformer encoder layers are expressed as:
\vspace{-0.2cm}
\[
E_0 = \text{concat}(E_{AV}, E_A, E_S, E_M) + PE + E_{\text{attr}},
\]
\vspace{-0.5cm}  % Adjust this value as needed
\[
E_{\text{enc}} = \text{Transformer}(E_0),
\]
where \(\text{Transformer}(E_0)\) applies layer normalization, multi-head attention, and feedforward networks sequentially, yielding the final encoded output \(E_{\text{enc}}\).

\subsection{Spatial Queries}

The framework employs a query-based decoding mechanism using distinct lateral and longitudinal queries to adapt to various driving scenarios. These queries guide decision-making by considering the vehicle's position and dynamics:
\vspace{-0.2cm}
\[
Q_0 = \text{MLP}(\text{concat}(Q_{\text{lat}}, Q_{\text{lon}})),
\]
where \(Q_{\text{lat}} \in \mathbb{R}^{N_R \times D}\) and \(Q_{\text{lon}} \in \mathbb{R}^{N_L \times D}\) represent lateral and longitudinal queries, respectively. The lateral queries, \(Q_{\text{lat}}\), are generated from a polyline encoder, which captures spatial information about drivable paths adjacent to the vehicle. These queries are essential for tasks that involve lateral movement, such as lane changes or obstacle navigation. The longitudinal queries, \(Q_{\text{lon}}\), are learnable embeddings that capture temporal dynamics, such as speed and acceleration. These are derived from the vehicle's historical and predicted driving data, enabling the model to anticipate and adjust to longitudinal changes in driving conditions.

\subsection{Historical Planning Embeddings}

As shown in Fig. \ref{fig:overview_structure}, the planning embedding at each timestep is obtained from the query-based decoder. The decoder consists of \(L_{dec}\) layers, each incorporating three types of attention mechanisms: lateral self-attention, longitudinal self-attention, and query-to-scene cross-attention. These mechanisms are mathematically defined as follows:
\vspace{-0.2cm}
\begin{equation}
\begin{aligned}
\label{fatorized_attn}
&Q'_{i-1} = \text{SelfAttn}(Q_{i-1}, \text{dim}=0), \\
&\hat{Q}_{i-1} = \text{SelfAttn}(Q'_{i-1}, \text{dim}=1), \\
&Q_i = \text{CrossAttn}(\hat{Q}_{i-1}, E_{enc}, E_{enc}).
\end{aligned}
\vspace{-0.2cm}
\end{equation}
Here, \(\text{SelfAttn}(X, \text{dim}=i)\) denotes self-attention applied across the \(i\)-th dimension of \(X\), and \(\text{CrossAttn}(Q,K,V)\) incorporates layer normalization, multi-head attention, and a feed-forward network. The output of the final layer is collected as the latent planning embedding for subsequent spatio-temporal aggregation.

\subsection{Spatio-Temporal Queries}

\begin{figure}[t]
  \centering
  \includegraphics[width=0.9\linewidth]{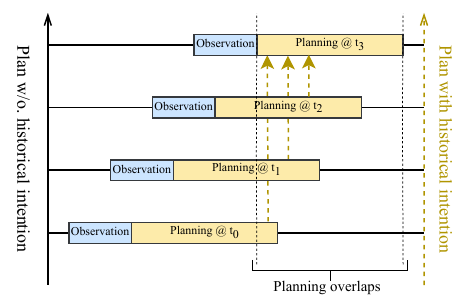}
  \vspace{-0.3cm}
   \caption{\textbf{Planning Stream Demonstration.} The diagram illustrates planning across consecutive frames, with dashed lines highlighting significant overlap that reveals the continuity and evolution of driving behaviors.}

   \label{fig:plan_comparison}
   \vspace{-0.6cm}
\end{figure}

As depicted in Figure \ref{fig:plan_comparison}, the overlapping time ranges within the planning streams provide valuable insights into the continuity and evolution of driving behaviors over time. These insights are critical for improving the robustness and effectiveness of planning algorithms by offering a deeper understanding of the vehicle's intentions across successive frames. To leverage this temporal overlap, we introduce a novel query-based decoder designed to aggregate and interpret spatial-temporal data. Specifically, we collect historical planning embeddings from various timesteps into a historical planning pool. This pool serves as a repository of past driving decisions and trajectories, which is then combined with reference queries that encapsulate the spatial context for the current driving scenario. The integration process combines the current spatial queries with the accumulated historical planning embeddings at the level of the reference line and mode:
\vspace{-0.5cm}
\[
\vspace{-0.2cm}
Q_{0}^{st} = \text{Projection}(\text{concat}(Q_{\text{0}}, \sum\limits_{i=-T_{H}}^{0}Q_{historical}^{\text{i}})),
\]
where \(Q_{\text{0}}\) represents the spatial queries at the current timestep, \(Q_{\text{historical}}\) denotes the aggregated historical planning embeddings over the past observation window \(T_{H}\), and \(Q_{0}^{st}\) is the fused spatio-temporal query used in the subsequent decoding process. This approach enables dynamic adjustments of the planning strategy based on both immediate and historical contexts, improving decision-making accuracy and the adaptability of autonomous systems in dynamic environments.

\subsubsection{Spatio-Temporal Decoder}

The final trajectory decoder structure closely mirrors that of the single-frame decoder described in Eq. \ref{fatorized_attn}. It consists of \(L_{dec}\) decoding layers, each featuring factorized attention mechanisms: lateral self-attention, longitudinal self-attention, and query-to-scene cross-attention. The key distinction is that the queries \(Q_{st}\) now integrate both spatial scale information from the reference line and temporal scale information from historical planning. The decoder’s output, \(Q_{dec}\), is used to predict the AV’s future trajectory points and their respective scores:
\vspace{-0.2cm}
\[
\begin{aligned}
&Q'_{i-1} = \text{SelfAttn}(Q_{i-1}, \text{dim}=0), \\
&\hat{Q}_{i-1} = \text{SelfAttn}(Q'_{i-1}, \text{dim}=1), \\
&Q_i = \text{CrossAttn}(\hat{Q}_{i-1}, E_{enc}, E_{enc}), \\
&\tau^{*} = \text{MLP}(Q_{dec}), \quad \pi^{*} = \text{MLP}(Q_{dec}).
\end{aligned}
\vspace{-0.2cm}
\]
Each predicted trajectory point consists of six channels: \([p_x, p_y, \cos \theta, \sin \theta, v_x, v_y]\). To handle cases without reference lines, an additional Multi-Layer Perceptron (MLP) head is introduced to directly decode a trajectory from the encoded AV features:
\vspace{-0.3cm}
\[
\vspace{-0.2cm}
\tau_{free} = \text{MLP}(E'_{AV}).
\]
\subsection{Training Loss}

To enhance the stability and efficiency of training, we structure the process into two primary phases. In the first phase, we train a standard Pluto planner \cite{pluto}, incorporating auxiliary losses: collision loss (penalizing collisions), contrastive imitation learning loss (CIL Loss, for mimicking expert driving), and prediction loss (for forecasting agent movements). These are collectively denoted as \(L_{pluto}\).

The second phase focuses on fine-tuning the Spatio-Temporal Decoder (STD). Here, we freeze the pre-trained planner's parameters and duplicate the planning embedding to handle multiple time steps. Training targets the STD and the trajectory\&score MLP, using smooth L1 loss for trajectory regression and cross-entropy loss for confidence classification. The ground truth trajectory, \(\tau_{gt}\), is projected onto a reference line, and the last segment guides the supervised target trajectory \(\hat{\tau}\). The applied losses are:
\[
L_{reg} = L1_{\text{smooth}}(\hat{\tau}, \tau_{gt}) + L1_{\text{smooth}}(\tau_{\text{free}}, \tau_{gt}),
\vspace{-0.2cm}
\]
\[
L_{cls} = \text{CrossEntropy}(\pi, \pi^*),
\]
where \(\pi^*\) is the one-hot encoded vector derived from \(\hat{\tau}\). The combined imitation loss, \(L_i\), integrates these components:
\[
\vspace{-0.2cm}
L_{\text{imitation}} = L_{reg} + L_{cls}.
\]
To penalize trajectories that violate dynamic constraints, we introduce a comfort loss \(L_{\text{comfort}}\) for each trajectory point. This loss ensures adherence to maximum dynamic parameters, such as longitudinal and lateral accelerations, yaw acceleration, yaw rate, longitudinal jerk, and jerk magnitude. The comfort loss is computed as:
\vspace{-0.2cm}
\[
L_{\text{comfort}} = \frac{1}{T_f} \sum_{t=1}^{T_f} \sum_{i \in \mathcal{P}} \max(0, |p_{i,t}| - p_{i}^{\max}),
\vspace{-0.2cm}
\]
where \(p_{i,t}\) is the measurement of each dynamic aspect at time \(t\), and \(p_{i}^{\max}\) is the maximum dynamic value.

%% file: sec/4_experiments.tex
\vspace{-0.2cm}
\section{Experiment}
\label{sec:exp}
\subsection{Experiment Setup}
\vspace{-0.1cm}
\noindent\textbf{Dataset.} We evaluate our method on the NuPlan dataset \cite{nuplan}, a large-scale closed-loop planning platform for autonomous driving, consisting of 1,300 hours of real-world driving data across 75 urban scenarios. It includes high-definition maps, sensor data, and labeled agent trajectories. The simulator runs each scenario for 15 seconds at 10 Hz.
% , where an LQR controller tracks the ego vehicle's trajectory generated by a planner. 

\noindent\textbf{Benchmark.} We evaluate our model on the Val14 benchmark \cite{PDM}, which includes 1,090 scenarios from 14 NuPlan challenge types, after excluding non-initializing reactive simulations. Agents are controlled in two modes: \textbf{Non-reactive}, following logged trajectories, and \textbf{Reactive}, using an Intelligent Driver Model (IDM) planner \cite{IDM}. Given the higher challenge posed by reactive simulations, which better reflect real-world conditions, the reactive closed-loop score is our primary metric.

\noindent\textbf{Metrcis.} NuPlan’s main evaluation metrics are the open-loop score (OLS), non-reactive closed-loop score (NR-Score), and reactive closed-loop score (R-Score). Since open-loop prediction shows minimal correlation with closed-loop planning, we focus on the following closed-loop metrics: \textbf{No ego at-fault collisions}: Recorded when the AV’s bounding box intersects with other agents or obstacles, excluding externally initiated collisions. \textbf{Drivable area compliance}: Ensures the AV stays within roadway boundaries. \textbf{Comfort}: Evaluates the AV’s comfort based on accelerations, jerks, and yaw rate. \textbf{Progress}: Compares the AV’s traveled distance to that of an expert driver, expressed as a percentage. \textbf{Speed limit compliance}: Verifies adherence to legal speed limits. \textbf{Driving direction compliance}: Monitors deviations from the correct direction, especially against traffic flow. A detailed description of these metrics is available in \cite{nuplan}.

\subsection{Implementation Details}
\vspace{-0.1cm}
We began experiments with the official Pluto checkpoint \cite{pluto}, using it as the backbone for the planner. Feature extraction focused on map elements and agents within a 120-meter radius, as per the NuPlan challenge. The planning horizon was set to 8 seconds, and the historical data horizon to 2 seconds for efficiency. Training was done on 8 NVIDIA A100 GPUs with a batch size of 64 for 5 epochs, using the AdamW optimizer with a weight decay of \(1 \times 10^{-4}\). The learning rate was gradually increased to \(1 \times 10^{-3}\) and decayed via a cosine schedule. All loss weights were set to 1.0, and total training time was approximately 6 hours.

\vspace{-0.1cm}
\section{Results and Discussion}
\label{sec:dis}
% In this section, we compare the performance of our model with current state-of-the-art (SOTA) methods, using both qualitative and quantitative analyses.  We examine factors that may impact performance, including the component modules, history length, and number of training epochs. Finally, we discuss the generalizability of our results across different model backbones and datasets.

\begin{table*}[htbp]
\centering
\vspace{-0.2cm}
\begin{tabularx}{\textwidth}{ll|ccccccc}
\toprule
\textbf{Type} & \textbf{Planner} & \textbf{Score} & \textbf{Collisions} & \textbf{Drivable} & \textbf{Progress} & \textbf{Comfort} & \textbf{Speed} & \textbf{Direction} \\

\midrule
\rowcolor{gray!30} \multicolumn{9}{c}{\textbf{Reactive Results}} \\
\midrule
Expert & Log-Replay & 81.24 & 86.38 & 98.07 & 98.99 & 99.27 & 96.47 & 99.13  \\
\midrule
\multirow{2}{*}{Rule-based} & IDM \cite{IDM} & 79.31 & 90.92 & 94.04 & 86.16 & 94.40 & 97.33 & 99.63\\
 & PDM-Closed \cite{PDM} & 93.20 & 98.67 & 99.63 & 90.48 & 95.50 & 99.83 & 100.00\\
\midrule
\multirow{5}{*}{Pure Learning} & PDM-Open \cite{PDM} & 54.86 & 81.47 & 87.89 & 69.86 & \textbf{99.54} & 97.72 & 97.57\\
 & GC-PGP \cite{GC-PGP} & 54.91 & 84.77 & 89.72 & 57.75 & 90.00 & \textbf{99.20} & 98.44 \\
 & RasterModel \cite{nuplan} & 64.66 & 87.11 & 83.11 & \underline{79.73} & 85.59 & 98.00 & 96.42\\
 & UrbanDriver \cite{UrbanDriver} & 64.87 & 82.16 & 90.83 & \textbf{86.42} & \underline{99.08} & 89.81 & 97.25\\
 & PlanTF \cite{plantf} & 77.07 & 95.69 & 97.06 & 76.72 & 93.94 & 98.67 & \underline{99.45}\\
 & PLUTO (w/o post.) \cite{pluto} & \underline{78.76} & \textbf{97.43} & \underline{97.89} & 75.08 & 96.33 & \underline{98.89} & \textbf{99.86}\\
 & LHPF† (ours) & \textbf{81.25} & \underline{95.82} & \textbf{99.63} & 78.59 & 96.42 & 98.60 & \underline{99.45}\\

\midrule
\rowcolor{gray!30} \multicolumn{9}{c}{\textbf{Non-reactive Results}} \\
\midrule
Expert & Log-Replay & 93.68 & 98.76 & 98.07 & 98.99 & 99.27 & 96.47 & 99.13 \\
\midrule
\multirow{2}{*}{Rule-based} & IDM \cite{IDM} & 79.31 & 90.92 & 94.04 & 86.16 & 94.40 & 97.33 & 99.63\\
 & PDM-Closed \cite{PDM} & 93.08 & 98.07 & 99.82 & 92.13 & 95.52 & 99.83 & 100.00\\
\midrule
\multirow{5}{*}{Pure Learning} & PDM-Open \cite{PDM} & 50.24 & 74.54 & 87.89 & 69.86 & \underline{99.54} & 97.72 & 97.57\\
 & GC-PGP \cite{GC-PGP} & 61.09 & 85.87 & 89.72 & 60.32 & 90.00 & \textbf{99.34} & 98.67\\
 & RasterModel \cite{nuplan} & 66.92 & 86.97 & 85.04 & 80.60 & 81.46 & 98.03 & 96.42\\
 & UrbanDriver \cite{UrbanDriver} & 67.72 & 85.60 & 90.83 & 80.83 & \textbf{100.00} & 91.58 & 98.49\\
 & PlanTF \cite{plantf} & 85.30 & 94.13 & 96.79 & 89.83 & 93.67 & 97.78 & 98.85\\
 & PLUTO (w/o post.) \cite{pluto} & \textbf{89.04} & \textbf{96.18} & \underline{98.53} & \underline{89.56} & 96.41 & \underline{98.13} & \textbf{99.49}\\
 & LHPF† (ours) & \underline{88.81} & \underline{94.77} & \textbf{99.72} & \textbf{90.90} & 96.88 & 97.77 & \underline{99.08}\\
\bottomrule
\end{tabularx}
\vspace{-0.2cm}
\caption{Closed-Loop Planning Results on the NuPlan Val14 Benchmark. All metrics are higher the better.}
\label{table:combined_results}
\vspace{-0.6cm}
\end{table*}

\subsection{Comparison with State of the Art}
We present a comparative analysis with other methods on the Val14 benchmark in Table \ref{table:combined_results}. Our LHPF method surpasses all other learning-based approaches in the total score of the Reactive closed-loop simulation, achieving an unprecedented score of 81.25. This achievement represents the first instance where a purely learning-based method exceeds expert performance, which was previously at 81.24. Additionally, LHPF excels in drivability, registering a score of 99.63. Compared to the baseline method, Pluto, the integration of historical planning information results in an improvement of the Progress metric from 78.59 to 79.21. This enhancement indicates that our model captures and executes the real driving intentions of the ego vehicle more precisely and adheres more closely to implicit driving behaviors. Nonetheless, there is a slight decline in collision-related metrics, likely attributable to the model's tendency to adopt more assertive maneuvers to achieve initial driving intentions. We plan to investigate this behavior further in subsequent ablation studies and will provide a detailed qualitative analysis.

\begin{figure*}[ht]
    \centering
\includegraphics[width=0.95\textwidth]{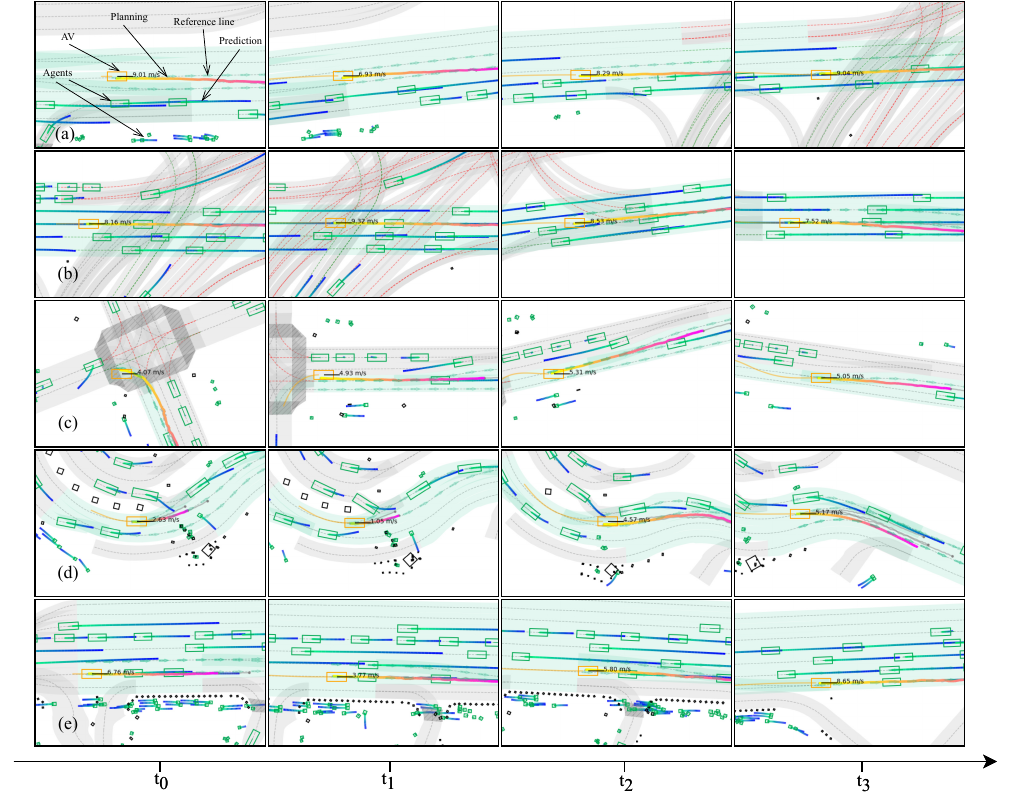}
   \vspace{-0.5cm}
    \caption{\textbf{Demonstration of Reactive Closed-Loop Simulation in Five Representative Scenarios.} The rectangles represent the vehicles, with the orange rectangle denoting the autonomous ego vehicle. Additional legends are provided in (a).}
    \label{fig:demo}
    \vspace{-0.6cm}
\end{figure*}

\subsection{Qualitative Results}

We present five representative scenarios in Fig. \ref{fig:demo}, used in closed-loop evaluation to demonstrate the performance. See the supplementary materials for more details.

\noindent\textbf{(a) An aggressive overtaking behavior.} The ego vehicle changes lanes to a neighboring lane at \( t_1 \), accelerates past vehicles in the original lane at \( t_2 \), and successfully completes the overtaking by \( t_3 \). This scenario highlights the ability of our method to make timely lane changes and execute aggressive overtaking maneuvers effectively.

\noindent\textbf{(b) A high-density traffic intersection scenario.} At time \( t_0 \), the autonomous vehicle is hindered by slow-moving traffic ahead. By \( t_1 \), it initiates a lane change maneuver and starts accelerating, successfully transitioning to an adjacent lane by \( t_3 \). This demonstrates how our model handles complex traffic interactions and efficiently maneuvers through dense environments.

\noindent\textbf{(c) A right-turn at an intersection.} The ego vehicle begins to turn right into the side road at \( t_0 \), fully entering it by \( t_1 \). At this moment, a vehicle from the opposite direction unexpectedly overtakes, forcing the ego vehicle to make a sharp right turn and realign back to the intended direction by \( t_3 \). This showcases the planner’s ability to react dynamically to sudden obstacles and adjust its trajectory smoothly.

\noindent\textbf{(d) A high-interaction roundabout scenario.} At \( t_1 \), the ego vehicle engages in a negotiation with a large vehicle attempting to merge into the main road. After yielding to let the large vehicle pass, it quickly accelerates in the neighboring lane at \( t_3 \), overtaking the large vehicle by \( t_4 \). This scenario highlights the planner’s capability to handle interactions with other road users and optimize vehicle trajectories in crowded environments.

\noindent\textbf{(e) An overtaking attempt.} Finally, we display an overtaking attempt by the ego vehicle. Due to the slow-moving vehicle ahead, the ego vehicle continuously decelerates between \( t_0 \) and \( t_1 \). At \( t_2 \), it attempts a left lane change to overtake but quickly abandons this maneuver due to a fast-approaching vehicle from the left rear, returning to the original lane at \( t_3 \) and continues to observe and wait for an overtaking opportunity. This showcases the flexible and adaptive nature of our method in dynamic driving scenarios.

\subsection{Results on CommonRoad}

\noindent\textbf{CommonRoad.} We evaluated the generalizability of our LHPF planner using the PGP backbone \cite{pgp} in the CommonRoad environment \cite{CommonRoad}. The planning embeddings from PGP's encoder were refined by retraining a latent variable-conditioned decoder (LVM), with no comfort loss applied to isolate the STD module’s generalizability. Following established protocols \cite{IR-STP, pop}, we trained on 239 scenarios from the scenario-factory and SUMO folders \cite{CommonRoad}, with a maximum of 8 scenarios per city. For closed-loop evaluation, we selected 54 interactive intersection scenarios, running simulations for 20 seconds to keep the ego vehicle within the map, while all agents (except the ego) were controlled by an intelligent driver model. The evaluation metrics are: completion distance (\textbf{DIST}), jerk cost (\textbf{JERK}), and valid front and side collisions (\textbf{FCT} and \textbf{SCT}, respectively). Baselines included \textbf{CA}, a search-based collision avoidance method, \textbf{NCA}, a search-based planner without collision avoidance, and \textbf{PGP}, a goal-conditioned lane graph traversal model.

\noindent\textbf{Results.} Table~\ref{table:commonroad} summarizes the closed-loop simulation results. L-PGP demonstrates balanced and robust performance. Notably, L-PGP achieves the higher DIST score (26.46), outperforming both the PGP baseline (25.18) and CA (19.21), supported by the integration of historical planning embeddings, which encourages assertive driving. Although this leads to a minor increase in JERK (45.67) compared to PGP (41.89), L-PGP substantially improves collision metrics, particularly in SCT, which decreases from 16 to 9, highlighting enhanced safety. In contrast, while N-CA achieves a low JERK (0.92), it incurs higher collision counts, underscoring trade-offs between smoothness and collision avoidance. Overall, these results validate the effectiveness of LHPF in optimizing both distance and safety performance in reactive closed-loop environment.

\begin{table}[t]
\centering
% \vspace{-0.2cm}

\vspace{-0.3cm}

\begin{tabularx}{\columnwidth}{l|ccc} % Use \columnwidth to span the entire column width
\toprule
Model & R-Score & Comfort & Progress\\
\midrule
Scratch                      & 78.76 & 96.33 & 75.08\\
\midrule
+ STD (Attn)                       & 77.76 & 95.69 & 74.13 \\
+ STD (Sum)                       & 80.53 & 93.67 & \textbf{80.42}\\
\midrule
+ STD (Attn) + Com.        & 79.17 & 95.96 & 77.43\\
\rowcolor{gray!30}   % Applying gray background to this row
+ STD (Sum) + Com.        & \textbf{81.25} & \textbf{96.42} & 78.59\\
\bottomrule
\end{tabularx}
\vspace{-0.2cm}
\caption{Ablation Study Results on NuPlan Val14 Benchmark}
\label{tab:ablation}
\vspace{-0.3cm}
\end{table}

\begin{table}[t]
\centering

% \vspace{-0.3cm}

\begin{threeparttable}
\begin{tabular}{l|cccc}
\toprule
Method & DIST$\uparrow$ & JERK$\downarrow$ & FCT$\downarrow$ & SCT$\downarrow$ \\
\midrule
CA\cite{IR-STP}      & 19.21 & 7.80 & 2 & 1 \\
N-CA\cite{IR-STP}    & 29.47 & 0.92 & 11 & 2 \\
\midrule
PGP\cite{pgp}        & 25.18 & \textbf{41.89} & 27 & 16 \\
\rowcolor{gray!30}   % Applying gray background to this row
L-PGP (ours)         & \textbf{26.46} & 45.67 & \textbf{26} & \textbf{9} \\
\bottomrule
\end{tabular}
\end{threeparttable}
\vspace{-0.2cm}
\caption{Closed-loop Simulation Results on Commonroad}
\label{table:commonroad}
\vspace{-0.6cm}
\end{table}

\subsection{Ablation Studies}

We conduct an ablation study to systematically examine the influence of various factors on the performance of the LHPF model. These factors include history planning length, fine-tuning epochs, fusion methods, and the incorporation of post-processing strategies. Unless otherwise noted, we use 10 scenarios per type from the NuPlan val14 dataset for these studies, given the time-consuming and impractical nature of implementing full simulations for all 1090 scenarios.

\noindent\textbf{Study of Each Component.} We initiated our analysis by comparing the performance of the base model against the enhanced model incorporating individual components. The scratch model, is essentially the Pluto model as detailed in Sec. \ref{sec:exp}. One notable observation is that adding a Spatio-Temporal decoder substantially encourages the planner to adopt a more aggressive strategy, resulting in a higher progress score (80.42). However, this comes at the cost of reduced comfort. Conversely, incorporating the comfort loss helps to balance performance between comfort and progress, culminating in the best R-score of 81.25.

% According to the nuPlan challenge's standard setup, the input observation length is 20 frames, resulting in historical planning embeddings spanning 1 to 20 frames. Figure \ref{fig:Epoch_score_history_score} demonstrates performance across four metrics at different historical intervals. The results show that longer intervals particularly improve the R-Score and Drivable metrics, which peak around the 10-frame interval before slightly declining. This pattern suggests an optimal planning interval that balances high performance across all metrics.

\begin{figure}[t]
  \centering
   \vspace{-0.5cm}
\includegraphics[width=0.85\linewidth]{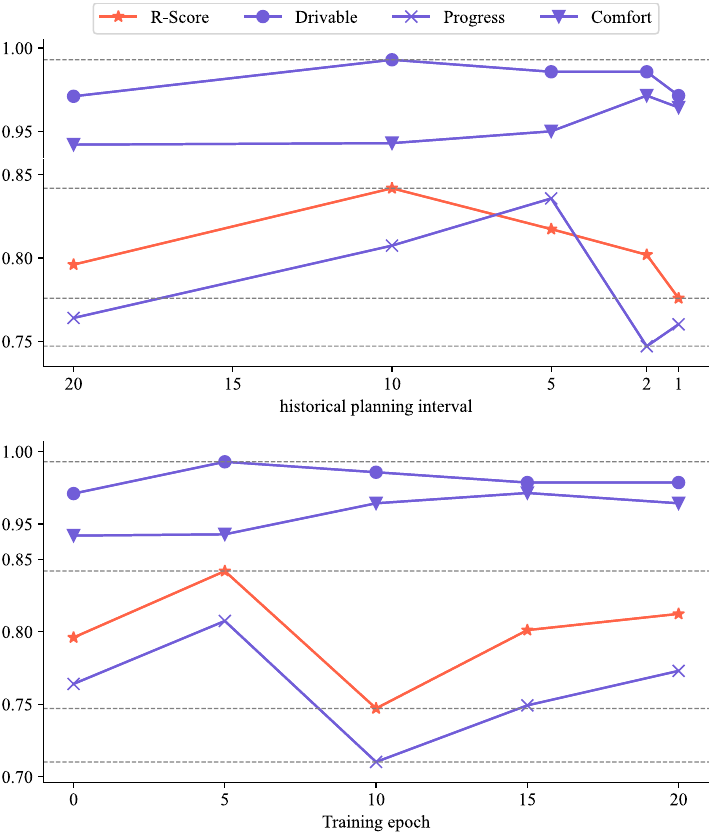}
\vspace{-0.2cm}
   \caption{Planning performance comparison with different historical planning intervals (upper) and training epochs (lower).}
   \label{fig:Epoch_score_history_score}
   \vspace{-0.6cm}
\end{figure}

\noindent\textbf{Study of Historical Planning Intervals.} According to the nuPlan challenge's standard setup, the input observation length is 20 frames, which we divided into different intervals. Figure \ref{fig:Epoch_score_history_score} shows performance across four metrics at various historical embedding intervals. When the interval is set to 20, our method degenerates to the baseline. Notably, at an interval of 10, where LHPF is first applied, all metrics improve. However, R-Score reaches its peak at interval 10, and further increasing the embedding density does not improve performance. In fact, at an interval of 1, R-Score drops below the baseline. This suggests that overly dense historical embeddings may introduce additional noise.

\noindent\textbf{Study of Training Epochs.} Given the sensitivity of our method’s performance to fine-tuning epochs, we systematically examined its relationship with planning performance, as shown in Figure~\ref{fig:Epoch_score_history_score}. Results indicate that optimal performance is achieved after approximately five epochs, further demonstrating that our method can be easily adapted to improve baseline models with only five epochs of fine-tuning for the Spatio-Temporal decoder. 

\noindent\textbf{Study of Fusion Approach.} Table~\ref{tab:ablation} presents a comparison of the performance between summation-based and attention-based fusion methods for two LHPF variants. Summation-based fusion consistently outperforms attention-based fusion across key metrics, with a higher R-score (80.53 vs. 77.76) and Progress (80.42 vs. 74.13), suggesting greater model reliability and planning effectiveness. When Comfort Loss is applied, this performance gap widens further; the Spatio-Temporal Decoder (Sum) with Comfort Loss configuration achieves the highest R-score (81.25) and Comfort score (96.42). These results indicate that summation-based fusion offers a more robust integration of historical planning features, making it the preferred approach for optimizing LHPF performance.

% \noindent\textbf{Study of Incorporation Post-Processing.} TODO. WAITING RESULTS. Add to the main table to illustrate the performance with post-processing. Waiting the results.

%% file: sec/5_conclusion.tex
\vspace{-0.15cm}
\section{Conclusion}
\vspace{-0.15cm}
This paper presents LHPF, a novel imitation learning planner designed to enhance autonomous driving safety by integrating historical planning information. Unlike traditional approaches that treat each plan in isolation, LHPF utilizes a historical intention aggregation module to maintain continuity and reduce errors. By incorporating a comfort auxiliary task, the method not only improves the human-likeness of driving behavior but also enhances planning accuracy. Our extensive experiments across diverse datasets and model backbones demonstrate that LHPF outperforms existing learning-based planners, showing promising potential for safer and more reliable autonomous driving.

%% file: sec/6_supplementary.tex
% \vspace{-0.cm}
% \section{1.Implementation Details}
% 参数表
% \section{3.Inference Latency}
% \section{4.Model Parameters}
% \section{5.Long term short term planning}
% \section{project page}
% \section{str2 pep eval}

\section{Supplementary Material}
\section*{A. Experiments on NuPlan}
\section*{a. Implementation Details}
% \subsection*{Input Representation and Scene Encoding}
\noindent\textbf{Dynamic Encoding.} Our model comprehensively encodes the state of each agent within the environment at time \(t\) using a vector \(s_t^i\) defined as \((p_t^i, \theta_t^i, v_t^i, b_t^i, I_t^i)\), where \(p\) and \(\theta\) denote position and heading, \(v\) represents velocity, \(b\) is the bounding box, and \(I\) is a binary observation mask. This sequence is transformed through differencing consecutive states to yield \(\hat{s}_t^i\). The resulting historical states are processed via a neighbor attention-based Feature Pyramid Network (FPN) to derive the agent embedding \(E_A\), with input dimensions \([N_A \times (T_H-1) \times 8]\) and output dimensions \([N_A \times D]\), where \(D\) is the dimension of the hidden layers.

\noindent\textbf{Static Encoding.} 
The MapEncoder processes map features by encoding each point in a polyline, capturing crucial data to define drivable areas. Each point's feature vector consists of relative positions and boundaries: \((p_i - p_0, p_i - p_{i-1}, p_i - p_{\text{left}}^i, p_i - p_{\text{right}}^i)\), with \(p_0\) marking the start of the polyline. These vectors form the input \(F_P \in \mathbb{R}^{N_P \times n_p \times 8}\), where \(N_P\) and \(n_p\) denote the number of polylines and points per polyline, respectively. This data undergoes transformation through a PointNet-like architecture to produce polyline embeddings \(E_M \in \mathbb{R}^{N_M \times D}\), with \(D=128\). Additional static features like traffic cones or barriers, represented by vectors \(o_i = (p_i, \theta_i, b_i)\), are processed by a two-layer MLP, generating features \(F_S \in \mathbb{R}^{N_S \times 5}\). These features are further encoded into embeddings \(E_S \in \mathbb{R}^{N_S \times D}\), enhancing the representation of static elements in the environment.

\noindent\textbf{Scene Encoding.} The encoded features \(E_A\), \(E_M\), and \(E_S\) are concatenated with the autonomous vehicle's state embedding \(E_{AV}\), forming an initial scene representation \(E_0\). This composite tensor is then enhanced with positional and semantic attribute embeddings, processed by \(L_{\text{enc}}\) transformer layers to yield the final scene encoding \(E_{\text{enc}}\), as shown below:
\begin{equation}
    E_0 = \text{concat}(E_{AV}, E_A, E_S, E_M) + PE + E_{\text{attr}},
\end{equation}
\begin{equation}
    E_{\text{enc}} = \text{Transformer}(E_0),
\end{equation}
where the transformer block applies sequential operations of layer normalization, multi-head attention, and feed-forward networks.
\\

\noindent\textbf{Historical Embedding Aggregation and Decoding.}
The decoding process integrates a query-based mechanism to model both lateral and longitudinal driving dynamics, adapting effectively to varying traffic conditions and driving scenarios. Queries are synthesized from encoded scene data and passed through a multi-layer perceptron (MLP) decoder, as well as self-attention and cross-attention mechanisms, to guide decision-making:

\begin{equation}
\begin{aligned}
% \label{fatorized_attn}
&Q_{i-1} = \text{MLP}(\text{concat}(Q_{\text{lat}}, Q_{\text{lon}})),\\
&Q'_{i-1} = \text{SelfAttn}(Q_{i-1}, \text{dim}=0), \\
&\hat{Q}_{i-1} = \text{SelfAttn}(Q'_{i-1}, \text{dim}=1), \\
&Q_i = \text{CrossAttn}(\hat{Q}_{i-1}, E_{enc}, E_{enc}),
\end{aligned}
\end{equation}

where \(Q_{\text{lat}}\) and \(Q_{\text{lon}}\) represent lateral and longitudinal queries, respectively, encapsulating critical aspects of the vehicle’s operational domain. The decoder comprises \(L_{dec}\) layers, each featuring factorized attention mechanisms: lateral self-attention, longitudinal self-attention, and query-to-scene cross-attention. The output queries, \(Q_{st}\), integrate spatial scale information from the reference line and temporal scale information from historical planning embeddings. This integration enables the decoder to predict the AV’s future trajectory points and their respective scores:

\[
\begin{aligned}
&\tau^{*} = \text{MLP}(Q_{dec}), \quad \pi^{*} = \text{MLP}(Q_{dec}).
\end{aligned}
\]
The final output of the decoder, \(Q_{dec}\), is used to generate the AV’s trajectory \(\tau^{*}\) and its corresponding confidence score \(\pi^{*}\), providing spatio-temporal consistency in planning. The inference process is aslo demonstrated in Algorithm \ref{al1}.

\begin{table}[htbp]
\centering
\begin{tabular}{l|c}
% \toprule
\textbf{Config} & \textbf{Value} \\
\midrule[\heavyrulewidth]
Optimizer & AdamW \\
Learning Rate & $1e-3$ \\
Weight Decay & $1e-4$ \\
% Learning Rate Schedule & Cosine \\
Batch Size & 64 \\
Training Epochs & 5 \\
Warmup Epochs & 3 \\
Historical interval & 10 \\
% Masking Ratio & [0.4, 0.5] \\
Loss Weight & [1.0, 1.0, 1.0] \\
Precision & 32.0 \\
Gradient Clip Value & 5.0 \\
% \bottomrule
\end{tabular}
\caption{Experiment Setting on NuPlan}
\label{table:exper_setting_nuplan}
\end{table}

% Table 7: Experiment Setting for Forecast-MAE Fine-Tuning
\begin{table}[ht]
\centering
\begin{tabular}{l|c}
% \toprule
\textbf{Config} & \textbf{Value} \\
\midrule[\heavyrulewidth]
Min Longitude Acceleration & -4.05 \\
Max Longitude Acceleration & 2.40 \\
Max Absolute Lateral Acceleration & 4.89 \\
Max Absolute Yaw Acceleration & 1.93 \\
Max Absolute Yaw Rate & 0.95 \\
Max Absolute Longitude Jerk & 4.13 \\
Max Absolute magnitute Jerk & 8.37 \\
% \bottomrule
\end{tabular}
\caption{Dynamics Constraints for Comfort Loss}
\label{table:Dynamics_constraints}
\end{table}

% Table 9: Runtime Evaluation 
\begin{table}[th]
\centering
\label{table:score_runtime_results}
\begin{tabular}{l|cc}
\toprule
\textbf{Planner} & \textbf{Score} & \textbf{Runtime (ms)} \\
\midrule
% \rowcolor{gray!30} \multicolumn{3}{c}{\textbf{Reactive Simulation}} \\
% \midrule
IDM \cite{IDM} & 79.31 & 27 \\
PDM-Closed \cite{PDM} & 93.20 & 90 \\
PDM-hybrid \cite{PDM} & 93.20 & 98 \\
PLUTO \cite{pluto} & 92.06 & 250 \\
\midrule
PDM-Open \cite{PDM} & 54.86 & \textbf{35} \\
GC-PGP \cite{GC-PGP} & 54.91 & 97 \\
RasterModel \cite{nuplan} & 64.66 & 73 \\
UrbanDriver \cite{UrbanDriver} & 64.87 & \underline{55} \\
PlanTF \cite{plantf} & 77.07 & 97\\
PLUTO (w/o post.) \cite{pluto} & \underline{78.76} & 102 \\
LHPF† (ours) & \textbf{81.25} & 129 \\
\bottomrule
\end{tabular}
\caption{R-Score and Runtime Results on the NuPlan benchmark. The upper section represents rule-based methods, while the lower section corresponds to pure learning-based methods.}
\label{table:runtime}
\end{table}

\section*{b. Experiment Setting}
We perform experiments using the official Pluto checkpoint \cite{pluto} as the foundational backbone of the planner. Feature extraction targets map elements and agents within a 120-meter radius, adhering to the NuPlan challenge specifications. The planning timeframe is set to 8 seconds, with 2 seconds of historical data for optimal efficiency. We run experiments on eight NVIDIA A100 GPUs with a batch size of 64 across five epochs. Detailed experimental settings are provided in Table \ref{table:exper_setting_nuplan}. We employ the AdamW optimizer with a weight decay of \(1 \times 10^{-4}\), and the learning rate progressively increases to \(1 \times 10^{-3}\) before following a cosine decay schedule. The historical embedding interval is set to 10. All loss weights are fixed at 1.0, and the total training duration is approximately 6 hours. Dynamic constraints used in the comfort loss are listed in Table \ref{table:Dynamics_constraints}.

\begin{algorithm}
\caption{Inference}
\begin{algorithmic}[1]
\setlength{\abovedisplayskip}{5pt} % Reduce space before equations
\setlength{\belowdisplayskip}{5pt} % Reduce space after equations

\State \textbf{Input:} $D, T_{H}, \psi$
\State \textbf{Output:} $\tau^{*}$
\State \textbf{Initialization:}
\State Initialize encoders, temporal aggregation module, and decoder
\State Load pre-trained weights if available
\State \textbf{Encoding:}
\State Extract features from input:
\[
E_{\text{enc}} = \text{Transformer}(D),
\]
\State \textbf{Historical Embedding Pool and Aggregation:}
\State Store historical planning embeddings in a pool $\mathcal{P}$
\State Fuse historical embeddings with the current frame:
\[
Q_{0}^{\text{st}} = \text{Projection}(\text{concat}(Q_{\text{0}}, \sum\limits_{i=-T_{H}}^{0}Q_{historical}^{\text{i}})),
\]
\State \textbf{Spatio-Temporal Decoding:}
\State Generate the future trajectory:
\[
\tau^{*} = \text{MLP}(\text{Spatio-Temporal Decoder}({E_{\text{enc}}, Q_{0}^{\text{st}}}))
\]
\State \textbf{Return:} $\tau^{*}$
\end{algorithmic}
\label{al1}
\end{algorithm}

\section*{c. Additional Results}

\begin{figure*}[ht]
    \centering
\includegraphics[width=1\textwidth]{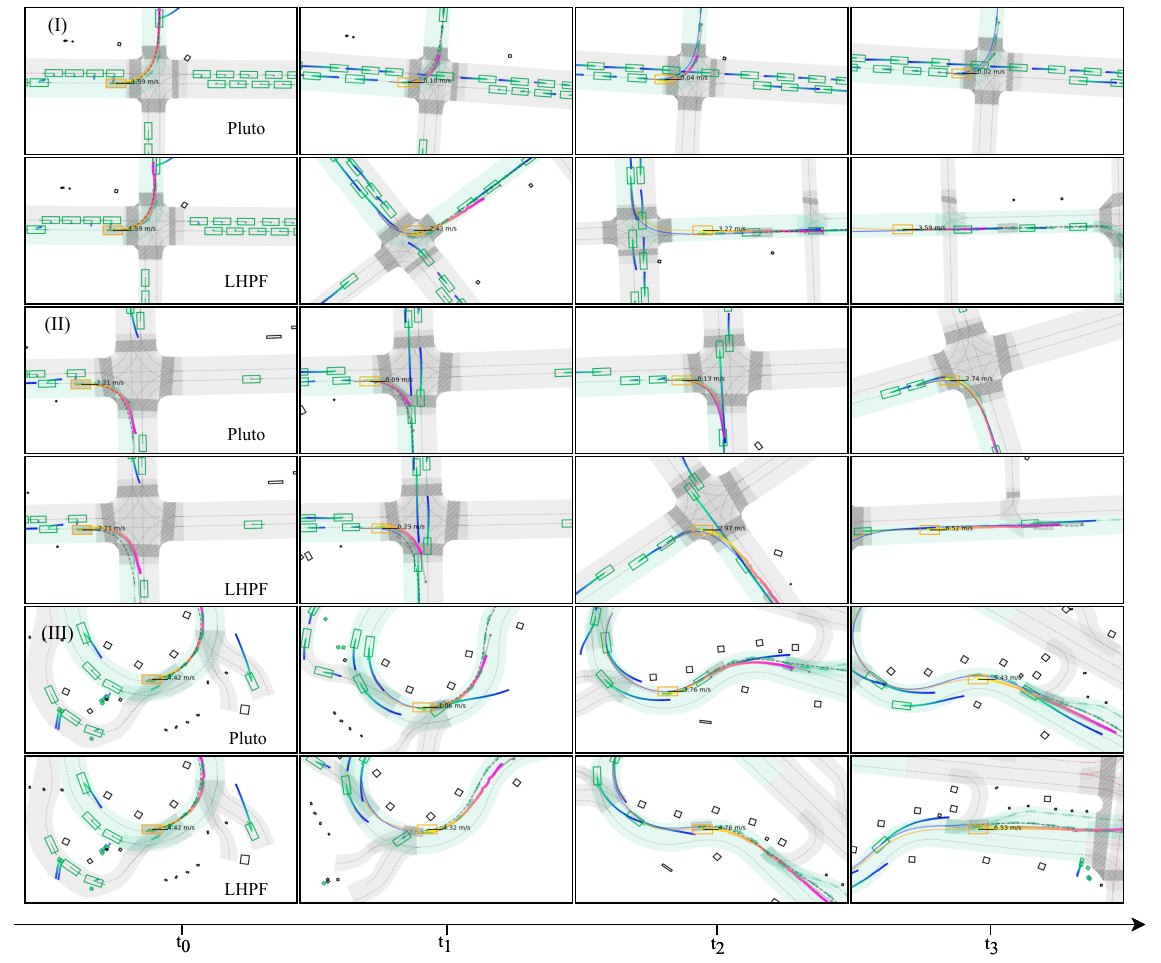}
    \caption{\textbf{More visualization comparisons of Reactive Closed-Loop Simulation on Nuplan V14 validation set.} The rectangles represent the vehicles, with the orange rectangle denoting the autonomous ego vehicle.}
    \label{fig:supp1}
\end{figure*}

\begin{figure*}[ht]
    \centering
\includegraphics[width=1\textwidth]{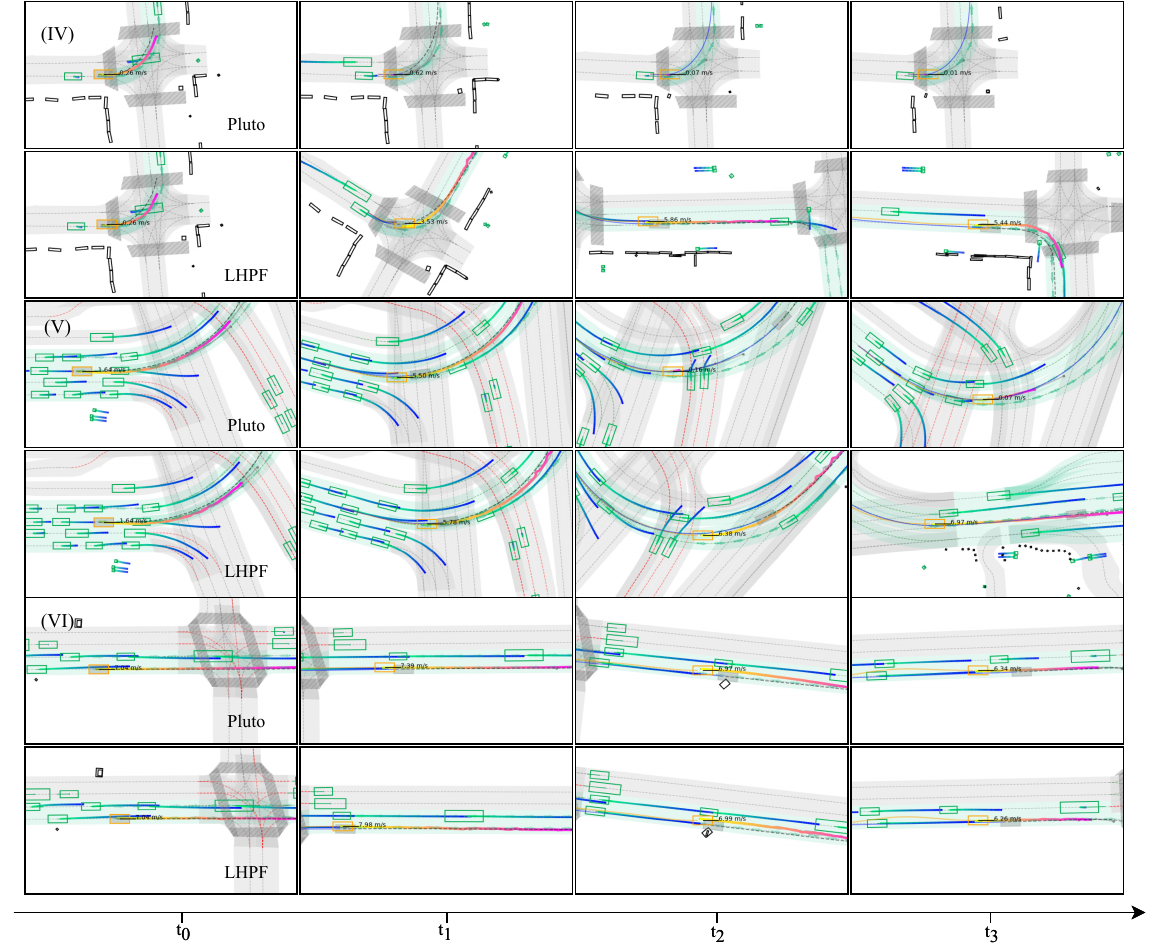}
    \caption{\textbf{More visualization comparisons of Reactive Closed-Loop Simulation on Nuplan V14 validation set.} The rectangles represent the vehicles, with the orange rectangle denoting the autonomous ego vehicle.}
    \label{fig:supp2}
\end{figure*}

\noindent\textbf{More Visual Results and Comparisons.} We compare the performance of LHPF with the SOTA baseline Pluto. Comparative visualization results are presented in Figures \ref{fig:supp1} and \ref{fig:supp2}. Rectangles represent vehicles, with the orange rectangle indicating the autonomous ego vehicle and green rectangles denoting dynamic agents such as vehicles, cyclists, and pedestrians. The ego plan is shown in yellow and pink, while agent predictions are depicted in green and blue. Reference lines are visualized as shaded arrows, and the ground truth trajectory appears as a gray shadow. Note that in reactive closed-loop simulation, \textbf{the ground truth is not the optimal solution} since observations change dynamically and it serves only as a recording of logged data. Panels (I) and (II) highlight the advantage of LHPF's more aggressive driving strategy, particularly in handling right-of-way scenarios during left and right turns at intersections. Panel (III) illustrates that while Pluto's conservative strategy often avoids risks, it fails to ensure safety in highly interactive scenarios, where LHPF demonstrates smoother interactions with surrounding vehicles. To better compare with the baseline, we have prepared some simulation visualization results in the visualization page \url{https://chantsss.github.io/LHPF/}.

In Panels (IV) and (V) of Figure \ref{fig:supp2}, LHPF showcases its ability to perform lane insertion maneuvers effectively in dense traffic environments. Lastly, Panel (VI) reveals an interesting observation: due to LHPF's aggressive driving style and typically higher speeds, when the perception module fails, the ego vehicle, despite immediately slowing down and yielding, still collides with a nearby vehicle. This observation explains why LHPF achieves better progress scores while showing a slight increase in collision metrics.

\noindent\textbf{Inference time experiments.} Table \ref{table:runtime} highlights the R-Score and runtime comparisons of LHPF against various baseline methods. The evaluations are conducted on a platform featuring an i9-12900KF CPU and an NVIDIA GeForce RTX 3080 Ti GPU. The upper section lists rule-based planners, such as IDM and PDM variations, while the lower section presents pure learning-based planners, including GC-PGP, RasterModel, and PlanTF. LHPF achieves the highest R-Score (81.25) among the pure learning-based planners, showcasing its superior planning accuracy. Although its runtime (129 ms) is slightly longer than simpler models like PDM-Open (35 ms) and UrbanDriver (55 ms), LHPF delivers a significant improvement in R-Score, ensuring a strong balance between accuracy and efficiency. This makes LHPF highly practical for real-world reactive simulation scenarios, where precise and responsive planning is essential.

\section*{B. Experiments on CommonRoad}
\section*{a. Implementation Details}
\noindent\textbf{Data Preparation.} CommonRoad~\cite{CommonRoad} serves as a benchmark for evaluating motion planners in a closed-loop environment on diverse road scenarios, which can be naturalistic, hand-crafted, or automatically generated. Following established protocols~\cite{IR-STP, pop}, we extracted interactive scenarios from the handcrafted, scenario-factory, and SUMO folders, generating 239 scenario simulations, each lasting 80 seconds, with a maximum of 8 scenarios per city. These scenarios were utilized to train the planner. Additionally, 54 highly interactive intersection scenarios were specifically chosen to evaluate the closed-loop performance of algorithms. The difference in the number of scenarios compared to \cite{IR-STP} may stem from updates to the CommonRoad dataset, resulting in variations despite using the same simulation procedure.

\noindent\textbf{Backbone.} We employ PGP~\cite{pgp} as the backbone for the planner. Given an HD map and the current states of agents, the model generates multiple future trajectories for the target agent, covering a prediction horizon of 6.0 seconds with 0.5-second intervals. PGP is a multi-modal prediction framework conditioned on lane graph traversal, comprising a graph encoder, a policy header, and a trajectory decoder. The graph encoder utilizes GRU layers to encode map context into node representations within a directed lane graph. Agent-node attention integrates information from surrounding agents into these nodes. Subsequently, GNN layers aggregate local context from neighboring nodes to refine the node encodings. Using these final node encodings, the policy header employs MLP scoring layers to learn a discrete policy for sampled graph traversals, identifying the most probable edges for the target agent to traverse. Finally, the trajectory decoder, conditioned on a latent variable and without incorporating comfort loss, predicts multiple potential future trajectories for the target agent.

\section*{b. Experiment Setting.}
The training details are summarized in Table \ref{exper_setting_commonroad}. The model is trained using the AdamW optimizer with a batch size of 96 on an NVIDIA GTX3090 GPU. The initial learning rate is set to \(2 \times 10^{-3}\) and decays by a factor of 0.6 every 50 epochs. The loss terms include \( \text{minADE}_5 \), \( \text{minADE}_{10} \), \( \text{MissRate}_5 \), \( \text{MissRate}_{10} \), and \(\pi_{\text{bc}}\). Specifically, ADE calculates the average Euclidean distance between predicted and ground truth trajectories, while the miss rate measures whether the deviation between the predicted trajectory and the ground truth exceeds a specified threshold. The subscript \(k\) denotes the top \(k\) predictions with the highest probability scores. The term \(\pi_{\text{bc}}\) represents the negative log-likelihood of the ground truth traversed edges under the learned policy. Loss weights are set to 1.0, 1.0, 1.0, 1.0, and 0.5, respectively. The PGP model is pre-trained for 100 epochs, followed by an additional 50 epochs to train the L-PGP model. After training, closed-loop experiments are conducted on selected scenarios using reactive simulations lasting 20 seconds. In these simulations, the ego vehicle planner can be CA, NCA, PGP, or L-PGP, while all other agents are controlled by an intelligent driver model.

% Table 8: Experiment Setting for SSL-Lanes
\begin{table}[htbp]
\centering
\begin{tabular}{l|c}
% \toprule
\textbf{Config} & \textbf{Value} \\
\midrule[\heavyrulewidth]
Optimizer & AdamW \\
Learning Rate & $2e-3$ \\
L.R. Scheduler & StepLR \\
L.R. Schedule Gamma & 0.6 at 50\\
Batch Size & 96 \\
Loss Weight & [1.0, 1.0, 1.0, 1.0, 0.5] \\
Pre-Training Epochs & 100 \\
Training Epochs & 50 \\
% Augmentation & None \\
% \bottomrule
\end{tabular}
\caption{Experiment Setting on Commonroad}
\label{exper_setting_commonroad}
\end{table}

\noindent\textbf{Evaluation Metrics.} To evaluate different algorithms and validate the effectiveness of our proposed LHPF, we employ the following closed-loop evaluation metrics:

\begin{itemize}
    \item \textbf{Completion Distance (DIST):} The average distance (in meters) traveled by the autonomous vehicle (AV) before stopping or the simulation ends in each scenario.
    \item \textbf{Jerk Cost (JERK):} The average jerk cost measures the trajectory smoothness. It is calculated as \((j^2 \cdot dt)\), where the time interval \(dt\) is set to 0.1 seconds.
    \item \textbf{Front Collision Times (FCT):} The total number of front collisions experienced by the AV across all selected scenarios.
    \item \textbf{Side Collision Times (SCT):} The total number of side collisions experienced by the AV across all selected scenarios.
\end{itemize}

